\documentclass{article}
\usepackage{spconf,amsmath,graphicx,cite}
\usepackage{url}

\title{Synthesising Expressiveness in Peking Opera via Duration Informed Attention Network}
\name{Yusong Wu\sthanks{Yusong Wu performed the work while at Tencent.}\textsuperscript{1}, Shengchen Li\textsuperscript{1}, Chengzhu Yu\textsuperscript{2}, Heng Lu\textsuperscript{2}, Chao Weng\textsuperscript{2}, Liqiang Zhang\textsuperscript{2},  Dong Yu\textsuperscript{2}}
\address{\textsuperscript{1}Beijing University of Posts and Telecommunications\\
\textsuperscript{2}Tencent AI Lab}

\begin{document}
\ninept
\maketitle
\begin{abstract}

This paper\footnote{Generated audio samples are available online\url{https://lukewys.github.io/files/ISMIR2019-audio-sample.html}.} presents a method that generates expressive singing voice of Peking opera. The synthesis of expressive opera singing usually requires pitch contours to be extracted as the training data, which relies on techniques and is not able to be manually labeled. With the Duration Informed Attention Network (DurIAN), this paper makes use of musical note instead of pitch contours for expressive opera singing synthesis. The proposed method enables human annotation being combined with automatic extracted features to be used as training data thus the proposed method gives extra flexibility in data collection for Peking opera singing synthesis. Comparing with the expressive singing voice of Peking opera synthesised by pitch contour based system, the proposed musical note based system produces comparable singing voice in Peking opera with expressiveness in various aspects.
\end{abstract}
\section{Introduction}
\label{sec:intro}

Taking the advantage of speech synthesis, the synthesises of singing voice, which can be taken as speech with more expressiveness, obtained success in recent decades~\cite{blaauw2017neural,Hono2018}. This paper proposes to take the advantage of speech synthesis further where a more expressive form of singing, opera singing, will be synthesised. Particularly in this paper, a method that synthesises Peking opera (or Jingju as a pronunciation translation) by Duration Informed Attention Network (DurIAN)~\cite{yu2019durian} is proposed. Moreover this paper proposes to use phoneme sequence together with music score as the input of DurIAN rather than traditional fundamental frequency ($f_0$) of singing voice, where the data availability can be potentially beneficial due to extra flexibility.

Although there are few works focusing on the synthesises of Peking opera, or more broadly, opera, the synthesise of singing voice has been researched since 1962 when Kelly and Lochbaum~\cite{lochbaum1962speech} used an acoustic tube model to synthesise singing voice with success. Recently several works~\cite{blaauw2017neural,yi2019singing,kaewtip2019enhanced,wada2018sequential,chandna2019wgansing} use deep neural networks to synthesise singing voice which, known as parametric systems, process fundamental frequency (or pitch contour, $f_0$) and harmonics features (or timbre) separately. Specially for synthesising Peking opera, such systems demand generating pitch contour before applying acoustic models. However, as there lacks systematic studies on the expressiveness of the singing voice in Peking opera, it is infeasible to generate singing voice of Peking opera with such parametric systems using intermediate features.

To avoid generating such intermediate features, an duration informed auto-regressive generation system is proposed. Current state-of-the-art end-to-end systems~\cite{shen2018natural,vasquez2019melnet,ping2018deep} in Text-To-Speech synthesis (TTS) generate speech voice from more basic inputs such as text and is capable of synthesising speech voices with emotion and prosody~\cite{shen2018natural,yu2019durian} without multiple complex processes being considered. However, such end-to-end approaches may suffer from alignment failure resulting in unpredictable artifacts with words skipped or repeated especially when generating unexpected texts, which will not be acceptable in synthesising opera singing voice.

To make the synthesised audio more perceptually indistinguishable, a state-of-the-art synthesis system named Duration Informed Attention Network (DurIAN)~\cite{yu2019durian} is used in the proposed system. DurIANs may overcome the problem of alignment failures~\cite{yu2019durian} in end-to-end systems and the problem of over-smoothing~\cite{blaauw2017neural} in parametric systems at the same time. In usual case, the pitch contour $f_0$ is expected to be provided as the input. For Peking opera, collecting the data of pitch contour is not an easy task as most Peking opera recordings have accompaniment music and it is almost impossible to label pitch contour manually. As a result, this paper proposes to use phoneme sequence together with music score to synthesis expressive Peking opera singing. Especially combined with music accompaniment, it is easier to obtain the phoneme sequence and musical note than an accurate pitch contour. Thus the proposed system attempts to generate spectrogram of synthesised acoustic signals with DurIAN by the input of phoneme sequence with music score with an auto-regressive decoder, effectively making the DurIAN maps the pitch variations caused by expressiveness in singing voice of Peking opera.

To demonstrate that the expressiveness in terms of pitch contours can be learned by DurIAN together with other links between note and spectrogram, two objective evaluation are performed to evaluate the performance of the proposed system. 

The following sections of this paper are organised in the following way. Firstly the databases to be used are introduced together with unique properties of Peking opera in terms of both expressiveness and acoustics. Then the essential changes on DurIANs to adopt the input of music notes on music score are described with experiment setups. The resulted spectrogram and pitch contour are compared to demonstrate the capacity of learning expressiveness for DurIANs followed by a quick discussion and conclusion.

\section{Data}
\label{sec:data}

In the proposed system, the phoneme of lyrics and the described notes in music are expected to be annotated. Although being popular domestically and intercontinentally for centuries, Peking opera has received little effort in research and synthesis, and not many datasets exist. One of the most commonly used dataset is ``Jingju a cappella singing" datasets~\cite{Gong2019,Gong2018,Gong2018b,Gong2018a} that contain over 7 hours of a cappella Peking opera recordings in a sample rate of 44.1kHz among which 1.71 hours recordings of 71 Peking opera singing have the phoneme annotated. The annotation of phoneme is adopted in a modified X-SAMPA (Extended Speech Assessment Methods Phonetic Alphabet) phoneme set\footnote{\url{https://github.com/MTG/jingjuPhonemeAnnotation}} containing 38 phonemes is used.

Besides the phoneme annotations, transcribed notes are also demanded in the proposed work, which is also provided by music transcription algorithm. The transcribed notes are essentially note-pitch and note-state estimated by pYIN algorithm~\cite{mauch2014pyin} where the note-pitch is rounded to integers and decoded using a Peking opera specific music language model~\cite{gong2016pitch} to make the boundaries of note clearer. The note-state consists of three states: attack, steady and silence.

\section{Methods}
\label{sec:method}

\begin{figure}[bht] 
 \centerline{
 \includegraphics[width=0.5\columnwidth]{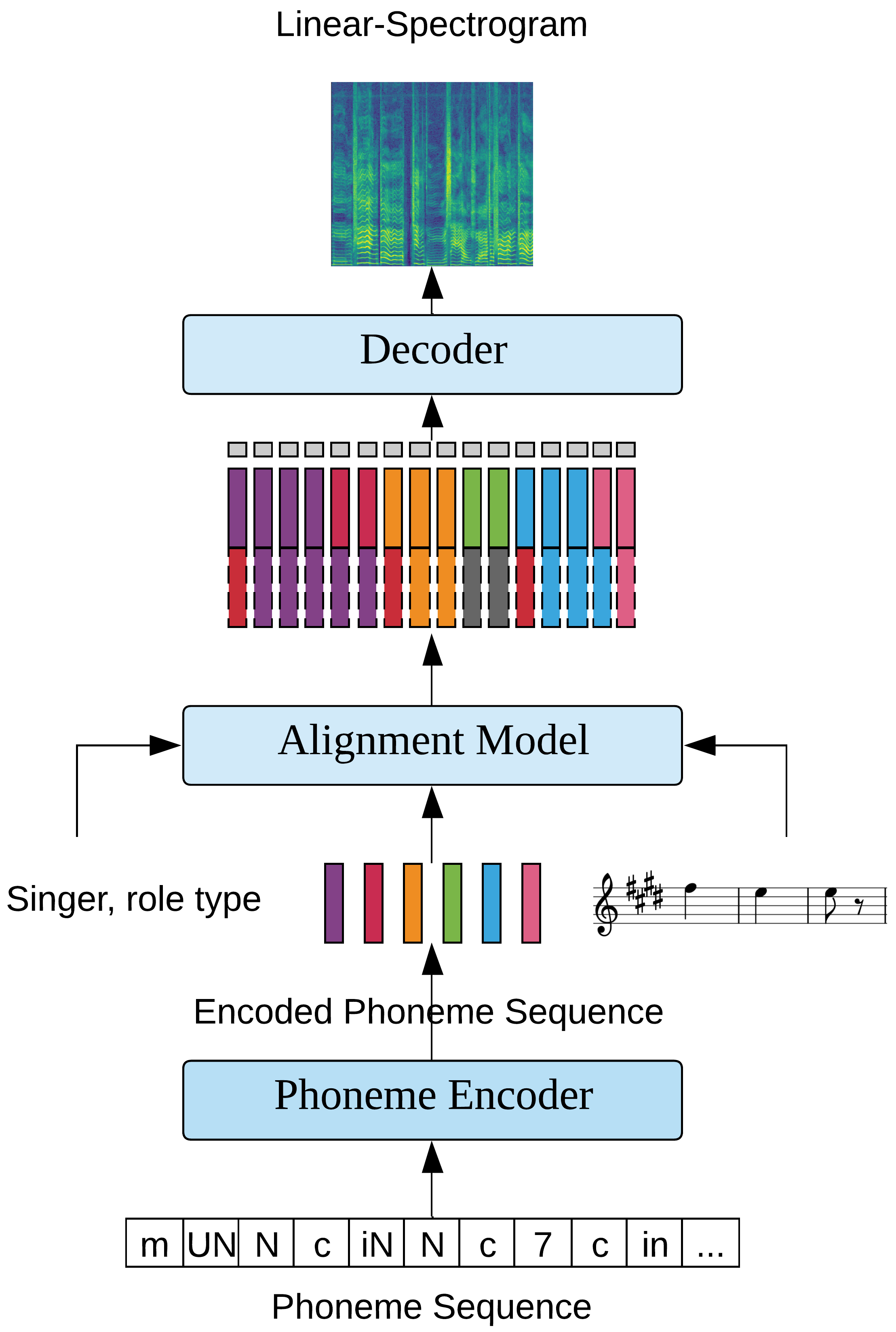}}
 \caption{The overview of our system. A phoneme sequence is encoded through a \textbf{phoneme encoder}, and aligned with high-level musical features in \textbf{alignment model}. Spectrogram is generated by applying an auto-regressive decoder to the aligned sequence.}
 \label{fig:model}
\end{figure}

To synthesise Peking opera singing, the proposed system firstly uses transcribed notes and phoneme sequence to generate the spectrogram of singing voice and the final acoustic signals are obtained by a selected synthesis system with the resulting spectrogram such as Griffin-Lim~\cite{griffin1984signal}. As a result, the expressiveness in singing voice is learned from the phoneme sequences together with the transcribed notes hence the key part of the proposed system is to generate expressiveness in the synthesised singing voice when the spectrogram is generated.

\subsection{Spectrogram Generation}

DurIAN for speech synthesis~\cite{yu2019durian} (\textit{DurIANs} in context) is able to provide extra robustness and efficiency on the synthesis process with better control~\cite{yu2019durian}. To generate expressiveness in singing voice, the proposed system selects DurIAN to generate spectrogram for Peking opera singing, which is represented as \textit{DurIANo} in the context.

DurIAN is an auto-regressive speech synthesis method, which generates spectrogram frames according to the temporal dependencies. In a DurIAN, the features of temporal dependency in phoneme sequences is modelled by a \textbf{phoneme encoder} and is aligned with acoustic signals where the resulted \textbf{alignment model} expand the encoded sequence according to the phoneme duration and concatenate with singing identity and musical features. The final spectrogram frames are generated by a \textbf{decoder} where the temporal dependencies in spectrogram and the aligned phoneme duration are modelled. With DurIAN, the proposed system is particularly good at avoiding the over-smoothing problem~\cite{yu2019durian} in singing voice synthesis and Fig.~\ref{fig:model} shows the overall architecture of the DurIAN used in the proposed system.

In general, the DurIANo makes little changes with DurIANs expect the parts related to expressiveness generation. Hence the description of DurIANo will emphasise on the differences between DurIANs and DurIANo in most cases.

\subsubsection{Phoneme Encoder}

The phoneme encoder in DurIANo makes little change with the DurIANs, which has a pre-net of two-layer fully connected neural network where a linear transformation is applied for better convergence. A CBHG module~\cite{wang2017tacotron} that consists of an 1D convolution bank with highway network and bidirectional GRU, is used to extract contextual representations from phoneme sequences. Unlike DurIANs,the skip gate in phoneme encoder is omitted as the X-SAMPA phoneme set used in our dataset does not contain prosody boundaries.

\subsubsection{Alignment Model}

\begin{figure}[bht] 
 \centerline{
 \includegraphics[width=\columnwidth]{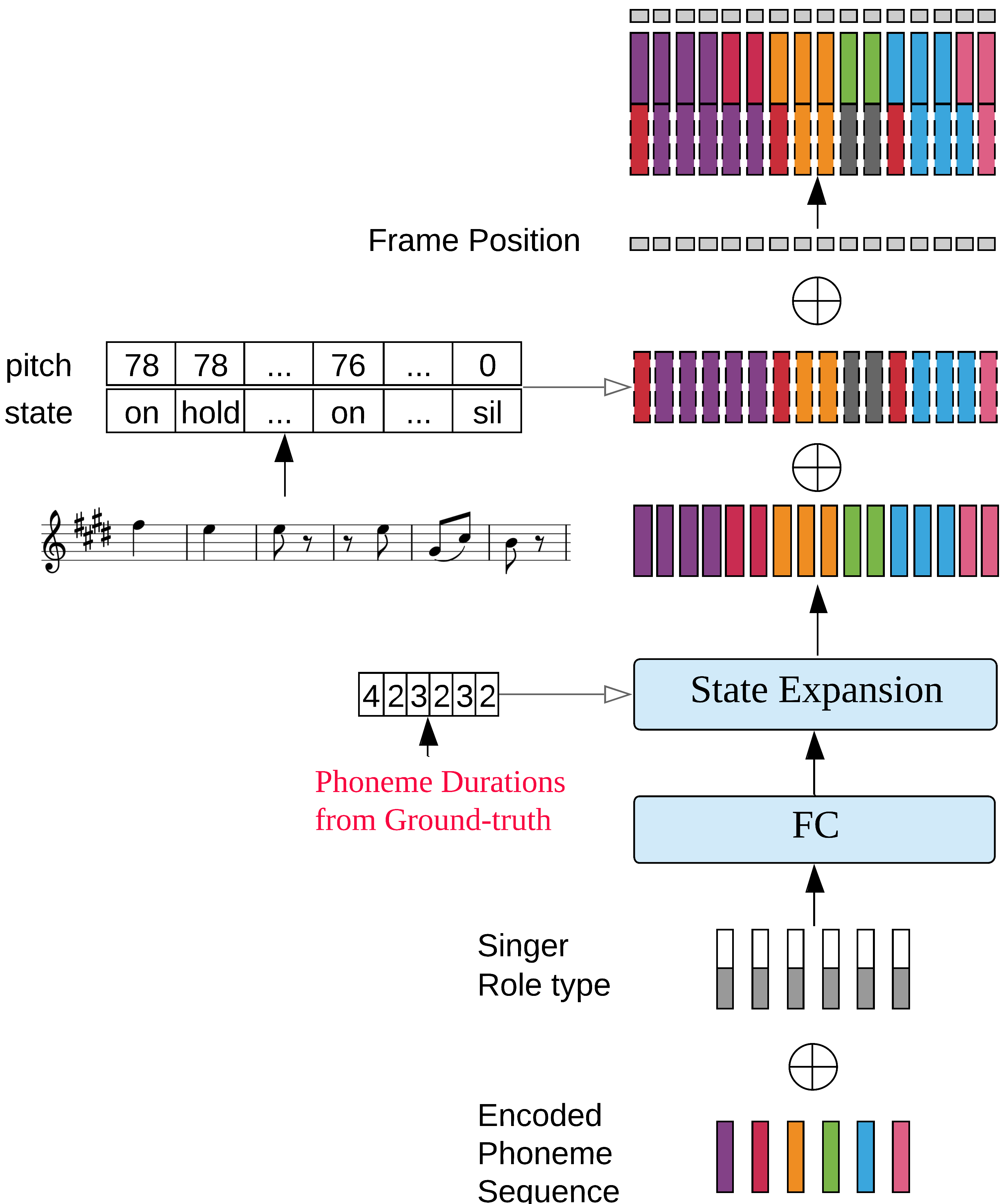}}
 \caption{The architecture of alignment model. First encoded phoneme sequence is encoded with singing identity by concatenation followed by a fully-connection. Then, the sequence is expanded according to the duration of each phoneme and aligned with the note-level musical features namely note-pitch. Last, frame positions are added.}
 \label{fig:model}
\end{figure}

In DurIANo, the expressiveness of singing is learned in this model, where the high-level music features such as transcribed note are aligned with the temporal features of phoneme sequence. As a result, the alignment model in DurIANo is different from DurIANs although the skeleton and the alignment algorithms of the model remains the same.

To align the transcribed notes and phoneme sequence in acoustic signals, the transcribed notes must be represented in a similar way with text to adopt DurIANs. As a result, the transcribed notes are represented by a description of music events rather than a sequence of note presence e.g. a A4 last for 5 frames are represented as $(A4,\text{`on'})$, $(A4,\text{`hold'})$, ..., $(sil,\text{`off'})$ rather than $(A4, A4, A4, A4, A4, sil)$ (where $sil$ represents silence). The transcribed notes is then aligned with the processed phoneme sequence in acoustic signals.

The annotated phoneme sequence in acoustic signals, which contains the phoneme duration of each phoneme in the acoustic signal, essentially represents expressiveness in the singing voice. DurIANo proposes to use fully connected networks to extracted features in phoneme sequences together with the singing identity, namely role type and the singer, then repeat each encoded phoneme vector in sequence according to its duration. The expanded states of frame-aligned phoneme with singer and role type information is then aligned with frame-aligned note event sequence by concatenating to frame-aligned sequence together where the expressiveness in singing voice is expected to be learned. After such process, a one-dimension scalar indicating the position of current input in the sequence in concatenated, allowing decoder to better learn the sequence dependencies.

\subsubsection{Decoder}
To generate synthesised singing voice, an auto-regressive decoder is proposed to generate spectrogram frame by making use of temporal information of existing spectrogram where the expressiveness in pitch is jointly generated with timber. Similar to the phoneme encoder, the spectrogram of acoustic signals are passed through a fully-connected pre-net for a non-linear transformation then a recurrent neural network with Gated Recurrent Units (GRU) combined with the frame-aligned feature sequence generates mel-spectrogram, which is needed as it contains highly compressed prosody, pitch and expressiveness information. Moreover, in DurIANo, two frames are generated each time by decoder RNN for faster and more stable convergence. Unlike DurIANs, the final spectrogram is generated by a CBHG layer rather than a post-net for improving performance of mapping and refining the decoder output to the final spectrogram as the CBHG can model bi-directional sequence dependencies. The resulted spectrogram is in a linear scale to fulfill the requirement of voice synthesis phrase.

\subsection{Voice Synthesis}
To generate audio from spectrogram, Griffin-Lim algorithm~\cite{griffin1984signal} is used, which transforms a linear-spectrogram into an acoustic signal by iteratively estimating the unknown phases. Specifically, the mean square error between the given spectrogram and the spectrogram of the predicted audio is iteratively decreased. Linear spectrogram is usually used as input rather than mel-spectrogram for more information is discarded in the latter , which would lead to the fail of spectrogram inversion. The use of Griffin-Lim is mainly for its promising of successful inversion and fastness in generation, as neural-network based vocoder such as WaveNet~\cite{Oord2016} or WaveRNN~\cite{kalchbrenner2018efficient} are found in the experiment to be hard to converge during training and very slow during inference.

\section{Experiments}
\label{sec:experiment}

\subsection{Experiment Setup}

The data used for training the proposed system contains 578 phrases where 17 phrases are taken as a separate validation set. Each training phrase is regarded as a training sample. During the training process, the number of mel-spectrogram is set to 80, the number of linear-spectrogram is set to 2049 with a hop length of 10ms and a window length of 50ms. The vectors of phoneme, singer and tole-type are all set to 256 as data collected from the dataset. The vectors of transcribed note ranges from C2 to C6 is set to 64 while the vectors of note state are set to 16. We train the decoder RNN and the final layer jointly using target of mel-spectrogram and linear-spectrogram which are computed by applying short-time Fourier transform (STFT) to the recording audio then mapped into the mel-scale and linear scale respectively. There is a $l1$ loss applied to RNN and post-CBHG layer outputs with the target of its respective spectrogram output in the proposed system during training process whereas a $l2$ regularisation has been applied for all trainable parameters. The system follows exact settings for training except that Griffin-Lim algorithm iterates for 60 times for singing voice generation that has a sampling rate of 44.1 kHz.

\subsection{Synthesis}

In synthesising Peking opera singing voice, the music note retrieved from the score is used instead of the note transcription result used in training to validate whether the proposed system can adapt to synthesising singing from score. The score used are published by Peking opera authority publisher, with different key and slightly different melody comparing to the corresponding recording.

The pitch of the note are manually aligned with phoneme boundaries. For the long phoneme (often the last phoneme of a word) which is usually supposed to be sang with a series of pitches, its duration is divided according to the note duration in the score. The initial consonant is set to silence to accord with note-transcription result used in training, and every single frame of which the onset of the note occurs is set to $on$. 

\subsection{Experiment References}

The resulted synthesised singing voice of Peking opera by the proposed system is compared with two types of references. The first type of references are original singing voice by Peking opera singers. The second type of references are generated by a baseline system which generates singing voice by pitch contours ($f_0$) extracted from original singing voice by SPTK tool box~\cite{sptk}. The baseline system uses same architecture, training data and parameters as the proposed system, except the note-pitch and note-state embedding vectors are replaced by a one-dimensional scalar indicating the fundamental frequency $f_0$. Such comparison demonstrates that the synthesised singing voice of Peking opera by proposed system is comparable with traditional methods depending on $f_0$ and thus the expressiveness of singing voice in Peking opera may be learned from high-level music features.

\section{Results}
\label{sec:results}
\begin{figure}[bht] 
 \centerline{
 \includegraphics[width=\columnwidth]{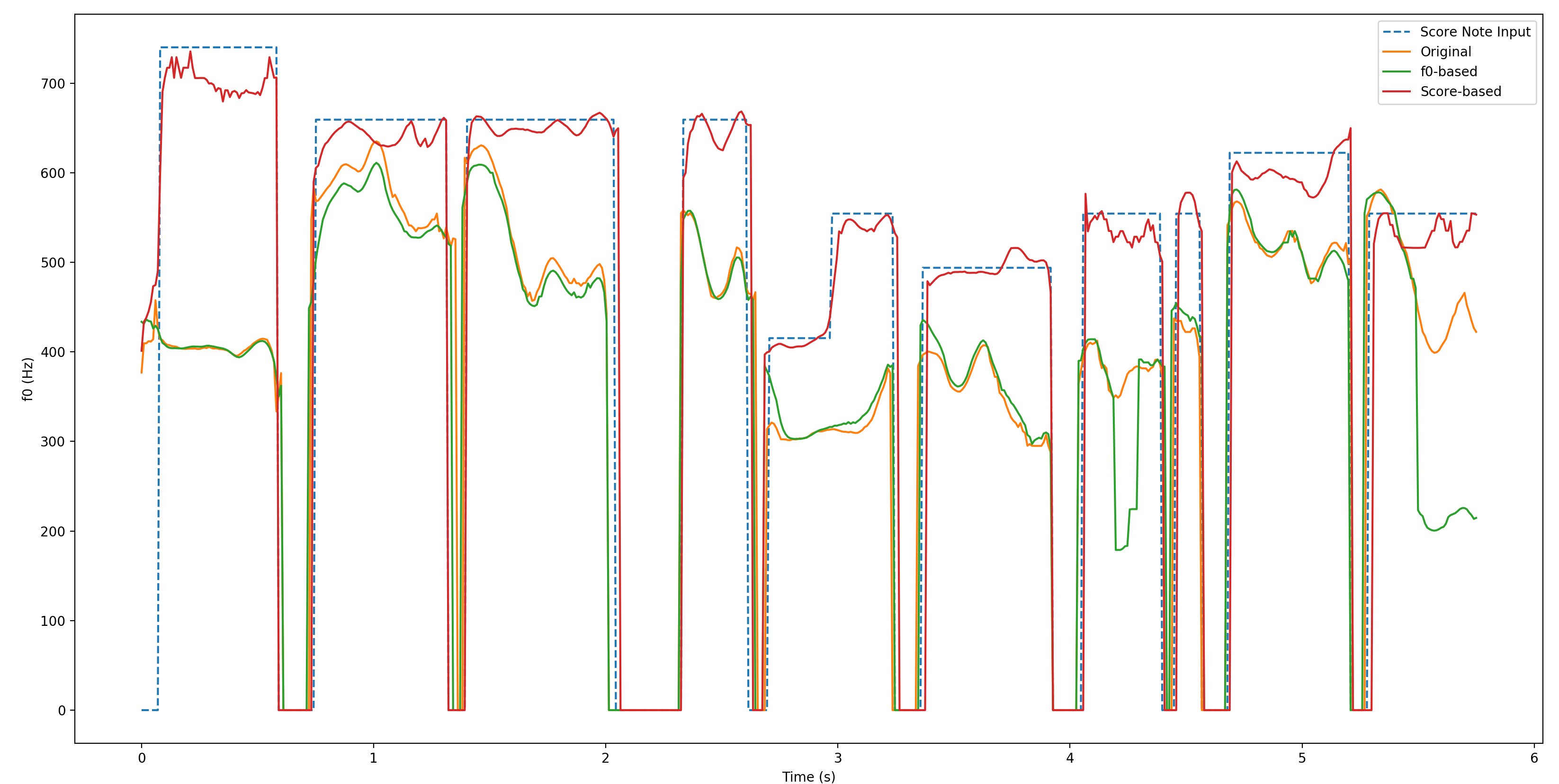}}
 \caption{The comparison of the $f_0$ of original recording and sample synthesized by $f_0$-based system and the proposed note-based system. The note pitch as input is not same as recording.}
 \label{fig:f0compare}
\end{figure}

In this experiment, eight phrases from a single piece of Peking opera singing is selected as validation set. This is to prevent the expressive style in validation set from being learned from the training set. Fig.~\ref{fig:f0compare} shows the pitch contour extracted from: 1) the synthesised singing voice by the proposed system and its input note; 2) the synthesised singing voice system generated from the reference system that learned Peking opera singing with pitch contour and 3) the human singing voice. Notice that the synthesised singing voice using score by the proposed system has different pitch with the recording.

To evaluate the performance of the proposed system and the reference system, two experiments are performed to evaluate the expressiveness in the synthesised singing voice. Firstly, the Pearson correlation coefficients are measured between the pitch contours of the synthesised singing voice to show the similarity of contours in terms of shape as shown in Table~\ref{tab:corrcoef}, which shows that the pitch contour in singing voice generated by the proposed system has higher correlation coefficients than the $f_0$ based system. These results reveal that the synthesised singing voice in the proposed system has higher correlation with the $f_0$ based system.

\begin{table}[htb]
    \centering
    \caption{The Pearson correlation coefficients resulted from pitch contours extracted from the synthesised singing voice by the proposed system (``score'') and the reference system (``$f_0$ based'') compared with the pitch contours extracted from human singing voice}
    \label{tab:corrcoef}
    \begin{tabular}{|c|c|c|c|}
    \hline 
       & original  & score & $f_0$ based \\ \hline
     original  & 1.0000 & 0.1276 & 0.0700 \\ \hline
     score  & 0.1276 & 1.0000 &  -0.0240 \\ \hline
     $f_0$ based  & 0.0700 & -0.0240 & 1.0000 \\ \hline
    \end{tabular}
\end{table}

Next a Gaussian model ($\mathcal{N}(\mu,\sigma)$) is used to model the distribution of pitch contour, where $\mu$ is the mean value of the distribution and $\sigma$ represents the standard deviation of the distribution. As the absolute pitch has little effect on expressiveness, the pitch contours are firstly standardised by setting the mean to 1. With expectation maximisation method, the pitch distribution in the pitch contours of original singing voice by singers is $\mathcal{N}(1.0000,0.1967)$ whereas the pitch contours generated by the proposed system and the reference systems are modelled by $\mathcal{N}(1.0000,0.1971)$ and $\mathcal{N}(1.0000,0.1766)$ respectively. This result demonstrates that the pitch contour generated by the proposed system has a higher similarity with the singing voice by performers compared with the traditional $f_0$ based system. The smaller values of standard deviation in pitch contour by the traditional $f_0$ based system suggests that a $f_0$ based system may produce less variable singing voice in terms of pitch, which can be effectively taken as the loss of expressiveness in pitch.

With the Gaussian model as the similarity measurement and the correlation coefficients as the measure of pitch contour shape, the objective measurements demonstrate that the proposed system outperforms the traditional $f_0$ system in terms of pitch contour. Moreover, as the pitch contours in the synthesised singing voice are more variable than the results of traditional $f_0$ system, this fact suggests expressiveness can be learned by the alignment of music score and annotated phoneme sequences in expressive performances, which could provide extra flexibility in collecting more Peking opera data for synthesis in future works.

\section{Conclusions}
\label{sec:conclusion}

This paper proposes a singing voice synthesis system for Peking opera. As the phoneme sequence can be manually annotated in recordings with accompanied music, the proposed system uses phoneme sequence together with music score instead of traditional pitch contour for extra flexibility in data collection. With duration informed attention network, the alignment between phoneme sequences in expressive Peking opera performance and music score enables the generation of Peking opera with expressiveness. The resulted pitch contour in the synthesised singing voice by proposed system show higher correlation and more similar pitch distribution in comparison of pitch contour based systems.


\nocite{black2014automatic}
\bibliographystyle{IEEEbib}
\bibliography{ref}

\end{document}